\documentclass[letterpaper]{article} 
\usepackage{aaai2027}  
\usepackage[hyphens]{url}  
\usepackage{graphicx} 
\usepackage{natbib}  
\usepackage{caption} 
\usepackage{algorithm}
\usepackage{algorithmic}
\usepackage{algorithm}
\usepackage{algorithmic}
\usepackage{amsmath,amssymb}
\usepackage{booktabs}
\usepackage{multirow}
\usepackage{newfloat}
\usepackage{listings}
\DeclareCaptionStyle{ruled}{labelfont=normalfont,labelsep=colon,strut=off} 
\floatstyle{ruled}
\newfloat{listing}{tb}{lst}{}
\floatname{listing}{Listing}

\usepackage{booktabs}

\title{PartMat: Material-Aware 3D Part Decomposition with a Single Global Latent}
\author{
    Guangming Fu\textsuperscript{\rm 1,2}\thanks{This work is done by Guangming Fu as an intern at Alibaba Group, supervised by Jin Song.}\equalcontrib,
    Jin Song\textsuperscript{\rm 2}\equalcontrib\corresponding, 
    Yiyun Fei\textsuperscript{\rm 2}, \\
    Guoqiu Li\textsuperscript{\rm 2}, 
    Ruigao Yang\textsuperscript{\rm 2}, 
    Jianan Jiang\textsuperscript{\rm 2}
}
\affiliations{
    \textsuperscript{\rm 1}Nankai University, China\\
    \textsuperscript{\rm 2}Alibaba Group, China

    \{fuguangming.fgm, songjin.song, yunhun.fyy, liguoqiu.lgq, ruigao.yrg, bingchen.jja\}@alibaba-inc.com
}

\begin{document}
\nocopyright
\maketitle

\begin{abstract}
Part-level 3D generation has recently attracted increasing attention for producing structured and editable 3D assets. However, existing methods typically decompose objects according to functional semantics rather than the editable material boundaries (e.g., fabric, wood, metal) required in practical 3D applications such as interior design. Additionally, current methods often generate parts independently, causing computational costs to scale linearly with the part count. To address these limitations, we present PartMat, an efficient material-aware 3D part decomposition pipeline that represents multi-part geometry with a single global latent. Given a reference image and a single whole-object geometry, PartMat decomposes the object into parts that follow material boundaries. First, we propose PartVAE to learn such a unified representation and decode all material parts in a single forward pass, thereby decoupling inference cost from the number of parts. Second, with this representation, a diffusion model is trained for part generation and refined via reinforcement learning for accurate material assignment and overlap suppression. Finally, to recover fine-grained geometric details, we introduce a sparse-voxel flow-matching model with part attention for geometry post-processing. Extensive experiments demonstrate that PartMat significantly outperforms existing baselines in material-aware decomposition accuracy and achieves comparable geometric quality, while maintaining efficient inference.
\end{abstract}

\section{Introduction}

\begin{figure}[t]
  \centering
  \includegraphics[width=\linewidth]{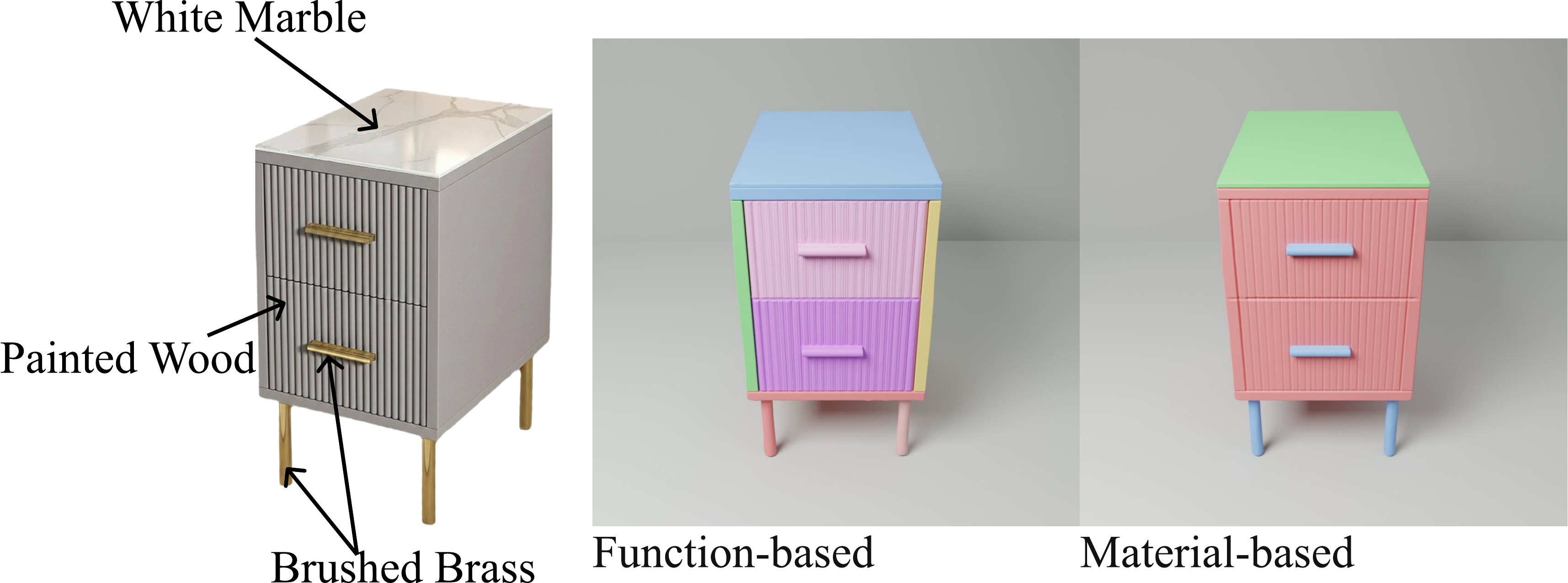}
  \caption{\textbf{Functional vs.\ material-aware decomposition.} Functional parts follow structural roles, whereas our components follow editable material assignments, such as marble, painted wood, and brass. A material component may contain multiple disconnected surface regions.}
  \label{fig:introduction}
\end{figure}

Editability has become an essential requirement in image-to-3D generation. Rather than treating a generated asset as a single, undivided mesh, downstream applications require it to be decomposed into components that can be selected, replaced, and assigned independent attributes. Existing methods decompose objects by geometric structure and functional semantics, such as chair backs, seats, and legs. However, this organization does not account for material boundaries that determine appearance and physical behavior. Separating fabric, wood, metal, and glass regions enables independent editing in interior design and the assignment of distinct properties, such as density, friction, and stiffness, for embodied simulation and manipulation planning. Motivated by these requirements, we study \emph{material-aware 3D part decomposition}: given a reference image and a single whole-object geometry mesh without part annotations, the goal is to decompose the input geometry into parts whose boundaries follow editable material assignments. As illustrated in Figure~\ref{fig:introduction}, our decomposition follows material boundaries rather than conventional function-oriented semantics. Here, ``material-aware'' refers to the geometric decomposition; predicting PBR parameters or physical properties remains a downstream task.

\begin{figure*}[t]
  \centering
  \includegraphics[width=\textwidth]{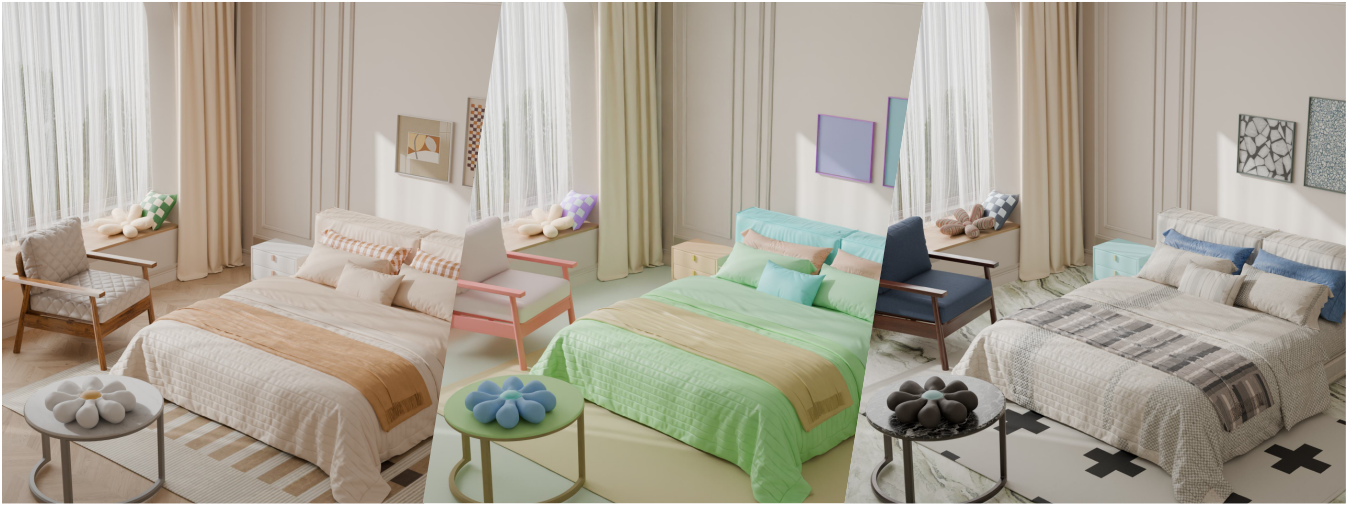}
  \caption{\textbf{Material-Aware Part Decomposition Results.} PartMat decomposes input 3D shapes into explicitly separated material-editable parts, each rendered in a distinct color for downstream PBR assignment and replacement. \textbf{Left:} A complex bedroom scene comprising various objects with rich materials. \textbf{Middle:} Material-aware decomposition of selected objects from the scene, as directly produced by our model. \textbf{Right:} Downstream applications enabled by our approach, demonstrating the decomposed assets seamlessly restyled with diverse PBR materials.}
  \label{fig:teaser}
\end{figure*}

Existing part-level methods mainly follow either segmentation-based or part-native pipelines. Segmentation-based methods decompose an input mesh primarily according to geometric structure or functional semantics~\cite{yang2025holopart,yan2026x}. Because geometry alone provides insufficient evidence for material boundaries, adapting these methods to material-aware decomposition requires first generating textures or material attributes on the mesh and then performing material segmentation and component reconstruction. This cascaded process lengthens the pipeline and propagates errors from appearance generation to the final decomposition.

Part-native methods instead learn multi-part representations directly, but often encode or generate each component separately~\cite{lin2026partcrafter,ding2025fullpart}. Despite avoiding explicit segmentation, these methods have two main limitations. First, their representation size and decoding cost grow with the number of components, leading to inefficient inference for objects with many material regions. Second, separate per-component representations can weaken global coordination across material regions. Moreover, latent flow matching supervises only the latent space and does not directly enforce consistency in the decoded geometry. This can cause small regions to be misassigned, disconnected regions of the same material to be split across different components, and generated parts to be duplicated or overlap with one another.

To address these challenges, we propose PartMat, a three-stage framework for image-guided material-aware decomposition. First, a global PartVAE encodes all material components into a single fixed-length latent and jointly decodes their signed-distance fields through a shared implicit backbone. Consequently, neither the latent size nor the number of decoder passes grows with the number of parts. Second, a conditional PartDiT generates this latent from the reference image and whole-object geometry. We then post-train PartDiT with direct-gradient reinforcement learning using differentiable rewards for component matching and overlap suppression. Third, to improve the geometric fidelity of the compactly decoded components, we introduce a high-resolution geometry post-processing stage. Figure~\ref{fig:teaser} demonstrates PartMat's material-aware decompositions in an indoor scene.
Experiments on approximately 300K material-annotated furniture and household objects show that PartMat achieves state-of-the-art material decomposition accuracy on our benchmark. 




\noindent\textbf{Contributions.} Our main contributions are as follows: 
\begin{itemize}
    \item We introduce PartVAE, to the best of our knowledge, which is the first VAE to represent multi-component geometry with a single global latent while decoupling the decoding latency from the number of parts.
    
    \item We propose PartMat, a three-stage framework that first employs direct-gradient reinforcement learning for precise material decomposition and overlap suppression, and applies a part-aware sparse flow refiner to restore high-resolution geometry with seamless component boundaries.
    
    \item Extensive experiments validate our method's superior performance in part-level decomposition and generation.
\end{itemize}
\section{Related Work}
\label{sec:related}

\subsection{3D Shape Representation}
Single-shape VAEs primarily use vector-set latents for implicit geometry~\cite{zhang20233dshape2vecset,dora,lai2026lattice} or sparse-voxel structures for high-resolution reconstruction~\cite{ren2024xcube,wu2026direct3d,xiang2026native}. Multi-part compression remains less explored. UniPart augments one holistic geometry latent with segmentation labels~\cite{he2026unipart}, whereas PartPacker packs non-contacting parts into two shared volume latents without dedicated per-part channels~\cite{partpacked}. PartMat instead compresses multiple explicit part geometries into one global latent.

\subsection{Object-Level Shape Generation}
Image-conditioned shape generation has progressed from transformer-based feed-forward reconstruction~\cite{hong2024lrm} to vector-set latent flow models trained at scale~\cite{li2025triposg,zhao2025hunyuan3d}. Recent systems adopt sparse-structured latents and spatial sparse attention for efficient high-resolution geometry~\cite{xiang2025structured,xiang2026native,wu2026direct3d}, while LATTICE combines voxel structure with set-based modeling~\cite{lai2026lattice}. Complementary to these latent shape representations, recent native mesh generators directly model explicit polygonal topology: SpaceMesh learns continuous manifold connectivity, MeshFlow generates continuous vertex-edge latents in parallel, and Nexus and LATO.2 factorize vertex and topology prediction~\cite{shen2024spacemesh,li2026meshflow,wang2026nexus,long2026lato2}. Despite improving geometric fidelity and scalability, these methods synthesize holistic assets without independently editable material parts. In contrast, PartMat natively decomposes shapes into independently editable material components, preserving high-frequency geometric details through a dedicated geometry refiner.

\subsection{Part Segmentation}

Classical geometry techniques~\cite{golovinskiy2008randomized,shapira2008consistent} and supervised point networks~\cite{qi2017pointnet,qi2017pointnetplusplus,wang2019dynamic,chen2019baenet,zhao2021pointtransformer} predict part labels but require hand-crafted cues or dense annotations. Recent methods lift 2D priors~\cite{kirillov2023segmentanything,abdelreheem2023satr,liu2023partslip,tang2024samesh,thai2024threebytwo,umam2024partdistill,zhong2024meshsegmenter}, learn open-world or native 3D features~\cite{yang2024sampart3d,liu2025partfield,ma2025p3sam}, or support language-grounded localization~\cite{ahmed2025kestrel}. Nevertheless, lifting can be inconsistent across views and occlusions, and these methods segment observed surfaces rather than recover complete material parts.

\subsection{Part-Level Shape Generation}
Early structured models use primitives or shape grammars, with PartNet providing fine-grained annotations~\cite{tulsiani2017primitives,li2017grass,mo2019partnet}. Part123 and PartGen segment views before reconstruction~\cite{liu2024part123,chen2025partgen}, while HoloPart and X-Part complete segmented 3D geometry~\cite{yang2025holopart,yan2026x}; both can propagate segmentation errors. PartCrafter and FullPart model per-part geometry with costs that scale with part count~\cite{lin2026partcrafter,ding2025fullpart}. BANG enables controlled decomposition through exploded dynamics~\cite{zhang2025bang}, whereas AutoPartGen generates variable-length part sequences autoregressively~\cite{chen2025autopartgen}. Most existing approaches organize parts according to geometric structure or functional semantics. Material boundaries need not follow these definitions: one functional part may contain multiple materials, and disconnected regions may share one editable material assignment. PartMat directly generates material-aware parts from a single global latent, with each part represented by an explicit SDF field.

\section{Method}

\begin{figure*}[t]
  \centering
  \includegraphics[width=\textwidth]{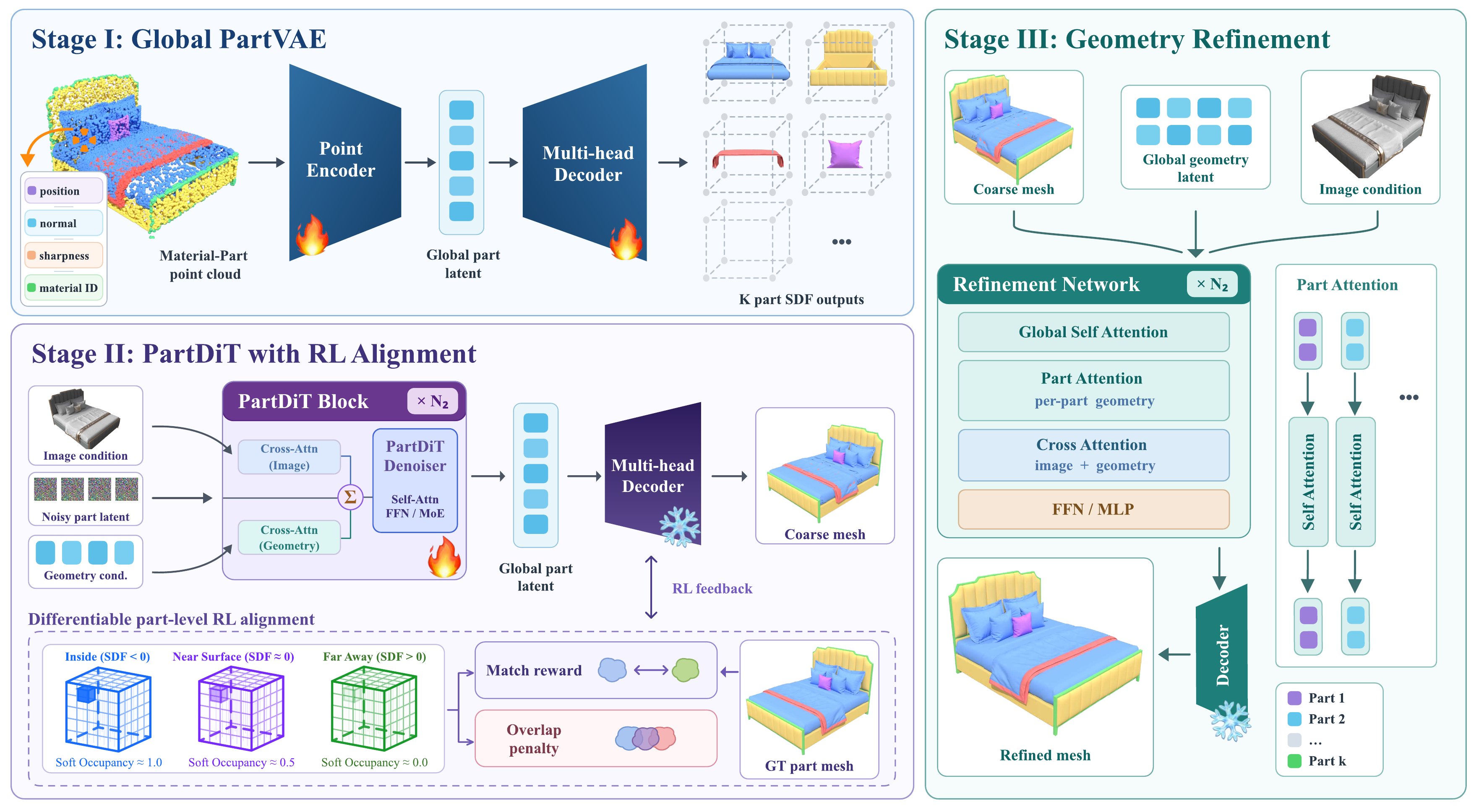}
  \caption{\textbf{PartMat overview.} Stage I encodes a part-aware point cloud into a global part latent and jointly decodes the component SDFs. Stage II uses separate cross-attention to the reference image and full-geometry conditions to generate this latent, and aligns the decoded fields with matching and overlap rewards. Stage III conditions on the coarse components, global geometry, and reference image to produce refined meshes through global, part, and cross-attention. Snowflakes and flames denote frozen and trainable modules, respectively.}
  \label{fig:pipeline}
\end{figure*}

The architecture of PartMat is illustrated in Figure~\ref{fig:pipeline}. Given a reference RGB image \(I\) and a single geometry mesh \(\mathcal M^g\), such as one generated by an image-to-3D model, PartMat produces a material-decomposable 3D asset
\(\mathcal O:=\{\mathcal M_k\}_{k=1}^{M}\). Each
\(\mathcal M_k=\{\mathbf V_k,\mathbf F_k\}\) is a separate mesh corresponding to one editable material component. All component meshes lie in the same object coordinate system as \(\mathcal M^g\), so they can be directly assembled into the complete object without additional transformations.

PartMat has three stages: PartVAE jointly encodes and decodes all material components; PartDiT predicts the global latent from the image and input geometry and is further optimized using differentiable SDF rewards; and a part-aware sparse flow model refines the components at high resolution.

\subsection{Stage I: PartVAE}
\label{sec:partvae}

\noindent\textbf{Architecture.}
PartVAE follows the variational encoder--decoder Transformer of Hunyuan3D 2.1 ShapeVAE~\cite{hunyuan3d21}: surface points are encoded into a sequence of latent tokens, from which spatial queries are decoded into SDF values. We adapt this architecture to material decomposition by adding material-part conditioning to the encoder and replacing the single SDF output with multiple part channels decoded from one global latent.

\noindent\textbf{Part-aware point encoder.}
Given an object with \(M\leq K\) annotated material parts, where \(K\) is the maximum number of part channels, we sample \(N\) points from the union of its part surfaces, including uniform surface samples and samples around sharp edges. The encoder input is
\begin{equation}
    \mathcal{P}=
    \{(\mathbf{x}_i,\mathbf{n}_i,m_i,s_i)\}_{i=1}^{N},
\end{equation}
where \(\mathbf{x}_i\), \(\mathbf{n}_i\), \(m_i\), and \(s_i\) denote the position, normal, material part ID, and sharp-edge indicator of point \(i\), respectively. Let \(\mathcal R=\{\mathbf r_\ell\}_{\ell=1}^{L}\) be voxel queries sampled from the occupied voxels of \(\mathcal P\). Our encoder Fourier-encodes the point and query coordinates, injects learned embeddings of \(m_i\) into the point features, and uses the voxel queries to aggregate \(\mathcal P\) through cross-attention and Transformer layers:
\begin{equation}
    \mathbf H
    =\operatorname{Encoder}_{\phi}(\mathcal P,\mathcal R).
    \label{eq:partvae_features}
\end{equation}
A linear pre-KL projection of \(\mathbf H\) predicts the Gaussian posterior parameters \(\boldsymbol\mu\) and \(\log\boldsymbol\sigma^2\). The global latent is then sampled as
\begin{equation}
    \mathbf z
    =\boldsymbol\mu+\boldsymbol\sigma\odot
      \boldsymbol\epsilon,\quad
    \boldsymbol\epsilon\sim\mathcal N(\mathbf 0,\mathbf I).
    \label{eq:partvae_encoder}
\end{equation}
where \(\mathbf z\in\mathbb R^{L\times d}\) is the global latent shared by all parts.

\noindent\textbf{Multi-channel SDF decoder.}
The decoder first maps \(\mathbf z\) back to the Transformer width with a post-KL projection. A latent Transformer then produces the decoded latent \(\mathbf z_{\mathrm{dec}}\). For each spatial query \(\mathbf q\), the point-query decoder cross-attends to \(\mathbf z_{\mathrm{dec}}\), and a \(K\)-channel linear head predicts all part SDFs:
\begin{equation}
    \begin{aligned}
    \mathbf h_{\mathbf q}
    &=D_{\mathrm{cross}}\!\left(
      \mathbf W_{\mathrm{squery}}\gamma(\mathbf q),
      \mathbf z_{\mathrm{dec}}\right),\\
    \mathbf F_{\psi}(\mathbf q;\mathbf z)
    &=[f_1(\mathbf q),\ldots,f_K(\mathbf q)]
      =\mathbf W_{\mathrm{sdf}}\mathbf h_{\mathbf q}+\mathbf b_{\mathrm{sdf}}.
    \end{aligned}
    \label{eq:part_sdf}
\end{equation}
The final projection produces one SDF channel per part. Thus, for a fixed \(K\), the decoding cost is independent of the actual part count \(M\). We extract each valid part mesh from its zero level set using Marching Cubes~\cite{lorensen1987marching}.

\noindent\textbf{Training objective.}
Let \(\mathcal Q=\{\mathbf q_j\}_{j=1}^{N_q}\) denote the sampled SDF query points. For each existing part \(k\), \(f_k(\mathbf q)\) and \(g_k(\mathbf q)\) are respectively the predicted and ground-truth SDFs at \(\mathbf q\in\mathcal Q\). We use \(\mathbf a\in\{0,1\}^K\) to mark existing parts and define \(\mathcal V=\{k\mid a_k=1\}\). Channels in \(\mathcal V\) are supervised by the reconstruction loss, while padding channels \(k\notin\mathcal V\) are kept away from the zero level set by the suppression loss. The overall objective is
\begin{equation}
    \mathcal L_{\mathrm{VAE}}=
    \mathcal L_{\mathrm{recon}}
    +\mathcal L_{\mathrm{suppress}}
    +\lambda_{\mathrm{KL}}\mathcal L_{\mathrm{KL}}.
    \label{eq:vae_objective}
\end{equation}
Let \(e_k(\mathbf q)=f_k(\mathbf q)-g_k(\mathbf q)\). The reconstruction term applies MSE and L1 losses to all valid parts:
\begin{equation}
    \mathcal L_{\mathrm{recon}}=
    \sum_{k\in\mathcal V}\sum_{\mathbf q\in\mathcal Q}
    \bigl[e_k(\mathbf q)^2+|e_k(\mathbf q)|\bigr].
    \label{eq:sdf_loss}
\end{equation}
For nonexistent parts, the suppression term constrains predictions to the negative interval
\([s_{\mathrm{lower}},s_{\mathrm{upper}}]\):
\begin{equation}
    \begin{aligned}
    \mathcal L_{\mathrm{suppress}}
    =\sum_{k\notin\mathcal V}\sum_{\mathbf q\in\mathcal Q}
    \bigl[
    &\operatorname{ReLU}
    (f_k(\mathbf q)-s_{\mathrm{upper}})^2\\
    +{}&
    \operatorname{ReLU}
    (s_{\mathrm{lower}}-f_k(\mathbf q))^2
    \bigr],
    \end{aligned}
    \label{eq:suppression}
\end{equation}
where \(s_{\mathrm{lower}}=-1.0\) and \(s_{\mathrm{upper}}=-0.1\). Finally,
\(\mathcal L_{\mathrm{KL}}=
D_{\mathrm{KL}}(q_{\phi}(\mathbf z\mid\mathcal P)
\Vert\mathcal N(\mathbf 0,\mathbf I))\).

\subsection{Stage II: PartDiT with RL Alignment}
\label{sec:rl_dit}

\noindent\textbf{PartDiT architecture.}
We freeze PartVAE and train a conditional generator in its latent space. PartDiT follows the Hunyuan3D 2.1 DiT architecture~\cite{hunyuan3d21}. DINOv2 encodes the reference image into tokens \(\mathbf c_I\)~\cite{oquab2023dinov2}. To obtain the full-geometry tokens \(\mathbf z^g\), we reuse the frozen PartVAE encoder and treat the complete input mesh \(\mathcal M^g\) as a single part. In each PartDiT block, the noisy part latent attends separately to \(\mathbf c_I\) and \(\mathbf z^g\) through two cross-attention layers.

PartDiT is trained with flow matching in the frozen PartVAE latent space. Let \(\mathbf z_1\) be the target part latent, \(\mathbf z_0\sim\mathcal N(\mathbf 0,\mathbf I)\), \(t\sim\mathcal U(0,1)\), and \(\mathbf z_t=(1-t)\mathbf z_0+t\mathbf z_1\). With \(\mathbf c=(\mathbf c_I,\mathbf z^g)\), the objective is
\begin{equation*}
    \mathcal L_{\mathrm{FM}}
    =\mathbb E\!\left[
    \left\|v_\theta(\mathbf z_t,t;\mathbf c)
    -(\mathbf z_1-\mathbf z_0)\right\|_2^2
    \right].
\end{equation*}

\noindent\textbf{Differentiable SDF reward.}
To improve the decomposition accuracy of the PartDiT, a natural choice is RL post-training. Applying Flow-GRPO~\cite{liu2026flow} with a mesh-IoU reward, however, requires explicitly decoding every sampled latent into component meshes before evaluation, resulting in high training cost. We instead compute a differentiable reward directly in the implicit SDF space of the frozen PartVAE.

For reward queries \(\mathbf q\in\mathcal Q_R\), the predicted and target fields are converted to soft occupancies
\(P_i(\mathbf q)=\sigma(f_i(\mathbf q)/\tau_s)\) and
\(T_j(\mathbf q)=\sigma(g_j(\mathbf q)/\tau_s)\).
Pairwise occupancy similarities \(S_{ij}\) are computed between all predicted and target channels. Because the predicted and target part channels may follow different orders, we use a differentiable soft assignment \(\pi_{ij}\) to match predicted channel \(i\) with target channel \(j\). Weighting \(S_{ij}\) by these assignments yields the order-invariant reward \(R_{\mathrm{match}}\). Meanwhile, \(R_{\mathrm{overlap}}\) penalizes simultaneous occupancy by different predicted components:
\begin{equation}
    \begin{aligned}
    R_{\mathrm{match}}
    &=\sum_{i,j}\pi_{ij}S_{ij},\\
    R_{\mathrm{overlap}}
    &=-\frac{1}{|\mathcal Q_R|\binom K2}
    \sum_{\mathbf q\in\mathcal Q_R}
    \sum_{1\leq i<j\leq K}P_i(\mathbf q)P_j(\mathbf q).
    \end{aligned}
\end{equation}
The combined SDF reward is
\begin{equation}
    R_{\mathrm{sdf}}
    =\lambda_{\mathrm{match}}R_{\mathrm{match}}
    +\lambda_{\mathrm{overlap}}R_{\mathrm{overlap}}.
    \label{eq:sdf_reward}
\end{equation}

\noindent\mbox{\textbf{Direct-gradient RL alignment.}}
Following LeapAlign~\cite{liang2026leapalign}, we directly backpropagate the differentiable SDF reward to PartDiT. Let \(\widehat{\mathbf z}\) denote a generated part latent; the frozen PartVAE decoder converts it into part SDFs for reward evaluation. The alignment objective is
\begin{equation}
    \mathcal L_{\mathrm{RL}}=
    \mathbb E\!\left[
    \max\bigl(0,c-R_{\mathrm{sdf}}(\widehat{\mathbf z})\bigr)
    \right],
    \label{eq:rl}
\end{equation}
where \(c\) is the reward margin.

\subsection{Stage III: Part-Aware Sparse Geometry Refinement}
\label{sec:geo_refine}

PartVAE provides coherent component geometry, but its compact latent and coarse SDF extraction can smooth thin structures and material interfaces. We refine all parts jointly with a sparse latent flow model conditioned on the coarse parts, the whole geometry, and the reference image.

\noindent\textbf{Sparse voxel.}
Diffusion on a dense \(512^3\) voxel grid is prohibitively expensive, whereas lowering the resolution loses surface detail; we therefore encode the coarse parts with a shape VAE and refine them on a fixed sparse voxel support~\cite{xiang2025structured}. The global input geometry and coarse parts are projected into this aligned sparse space to form the condition \(\mathbf U\). Coordinate construction and feature lookup are detailed in the supplementary material.

\noindent\textbf{Part attention.}
The refinement transformer must preserve part identity without processing every part in isolation. Let \(\mathbf X_k\) denote tokens with component index \(k\). We define within-part attention as
\begin{equation}
    \operatorname{PartAttn}(\mathbf X)=
    \operatorname{concat}_{1:k}
    \operatorname{SelfAttn}(\mathbf X_k).
    \label{eq:part_attention}
\end{equation}
Each refinement block interleaves full self-attention over
\(\mathbf X\), two part-attention layers, and cross-attention to the image. Part attention prevents features from losing their slot identity; full attention communicates object-level context across adjacent material interfaces.

\noindent\textbf{Sparse latent flow refinement.}
Let \(\mathbf X_1\) be the target features on the sparse structure
\(\widetilde{\mathcal C}\) and
\(\mathbf X_0\sim\mathcal N(\mathbf 0,\mathbf I)\). Following the rectified flow matching paradigm, we construct a linear interpolation path
\(\mathbf X_t=(1-t)\mathbf X_0+t\mathbf X_1\) and train the part-aware sparse transformer with
\begin{equation}
    \mathcal L_{\mathrm{HR}}=
    \mathbb E\!\left[
    \left\|H_{\omega}(\mathbf X_t,t,\mathbf c_I,\mathbf U)
    -(\mathbf X_1-\mathbf X_0)\right\|_2^2
    \right].
    \label{eq:hr_loss}
\end{equation}
At inference, we integrate the sparse flow on the fixed support and decode the sampled features. The resulting meshes remain in the shared object coordinate system and inherit the material slots produced by PartDiT.

\section{Experiments}

\subsection{Experimental Setup}
\label{sec:exp_setup}

\noindent\textbf{Training Data.}
We train PartMat on approximately 300K material-aware furniture and household-object shapes. Component labels are derived directly from the material slots assigned to mesh faces, with each slot defining one material component. Detailed preprocessing and sampling configurations are provided in the supplementary material.

\noindent\textbf{Baselines and Metrics.}
For VAE reconstruction, we compare against \textbf{Hunyuan2.1}~\cite{hunyuan3d21}, \textbf{TRELLIS2}~\cite{xiang2026native}, \textbf{PartCrafter}~\cite{lin2026partcrafter}, and \textbf{PartPacker}~\cite{partpacked}. All methods are evaluated at $512^3$ resolution. We report Chamfer Distance (CD), F1-Score at threshold 0.01 (F1@0.01), average latent token count, and VAE decode latency for 1, 16, and 32 components.
For image-conditioned generation, we compare against \textbf{X-Part}~\cite{yan2026x}, \textbf{PartCrafter}~\cite{lin2026partcrafter}, \textbf{PartPacker}~\cite{partpacked}, \textbf{HoloPart}~\cite{yang2025holopart}, and \textbf{OmniPart}~\cite{omnipart}. We report CD and F1@0.01 for whole-geometry fidelity and Sem-IoU after optimal component matching for material decomposition accuracy. Further evaluation details are deferred to the supplementary material.

\begin{figure}[t]
\centering
\includegraphics[width=\linewidth]{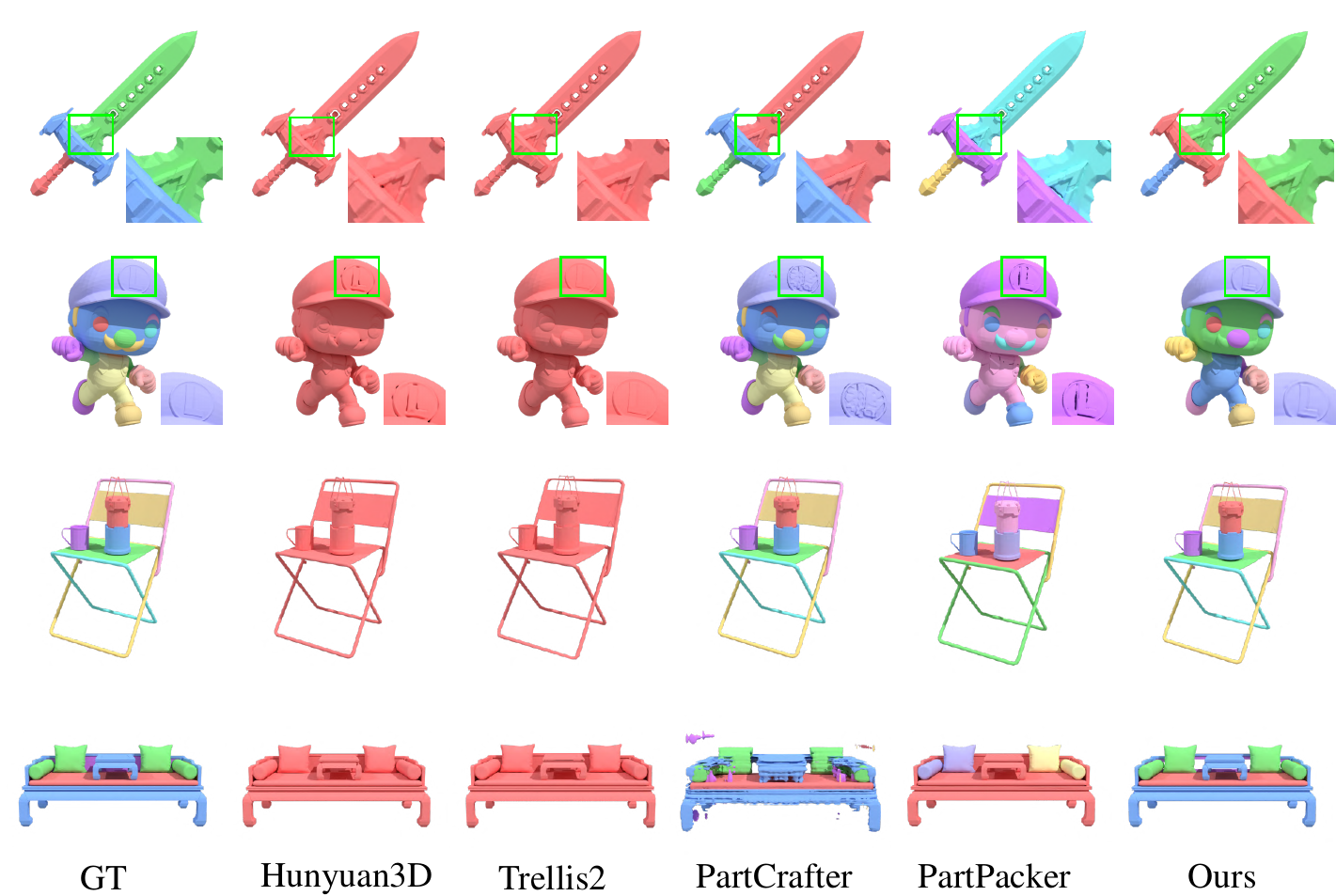}
\caption{Qualitative comparison of material-aware 3D reconstruction. PartVAE recovers complete geometry while preserving editable material-component labels, shown with distinct colors.}
\label{fig:vae_results}
\end{figure}


\begin{table}[t]
\centering

{
\scriptsize
\setlength{\tabcolsep}{2pt}
\resizebox{\linewidth}{!}{%
\begin{tabular}{lccc}
\toprule
Method & CD($\times\!10^{4}$)\, $\downarrow$ & F1@0.01 $\uparrow$ & Sem-IoU $\uparrow$ \\
\midrule
X-Part~\cite{yan2026x}               & 5.13 & 55.40 & 43.00  \\
PartCrafter~\cite{lin2026partcrafter}  & 12.32 & 10.27 & 15.11 \\
PartPacker~\cite{partpacked}        & 6.06 & 38.77 & 30.62 \\
HoloPart~\cite{yang2025holopart}        & 5.78 & 55.31 & 32.47 \\
OmniPart~\cite{omnipart}        & 5.60 & 30.38 & 29.75 \\
\midrule
PartMat & 4.90 & 68.23 & 46.92 \\
PartMat w/ RL & 3.67 & 67.45 & \textbf{50.51} \\
PartMat w/ geometry refine & \textbf{2.83} & \textbf{69.59} & 49.19 \\
\bottomrule
\end{tabular}
}
} 

\caption{Image-to-3D material-aware component generation comparison. Our PartMat significantly outperforms other methods on all metrics. Best results are marked in bold font.}
\label{tab:dit_comparison}

\end{table}

\noindent\textbf{Implementation Details.}
\label{sec:impl}
PartVAE follows Hunyuan3D 2.1-VAE~\cite{hunyuan3d21} and jointly decodes $K{=}32$ material-component SDF channels. PartDiT uses the Hunyuan3D 2.1 DiT backbone, and the geometry refiner augments TRELLIS2~\cite{xiang2026native} with part attention. Complete training schedules and hyperparameters are provided in the supplementary material. All models are trained on NVIDIA H20 GPUs, while decoding latency is measured on a single NVIDIA RTX 3090 GPU.

\begin{figure*}[ht]
\centering
\includegraphics[width=\linewidth]{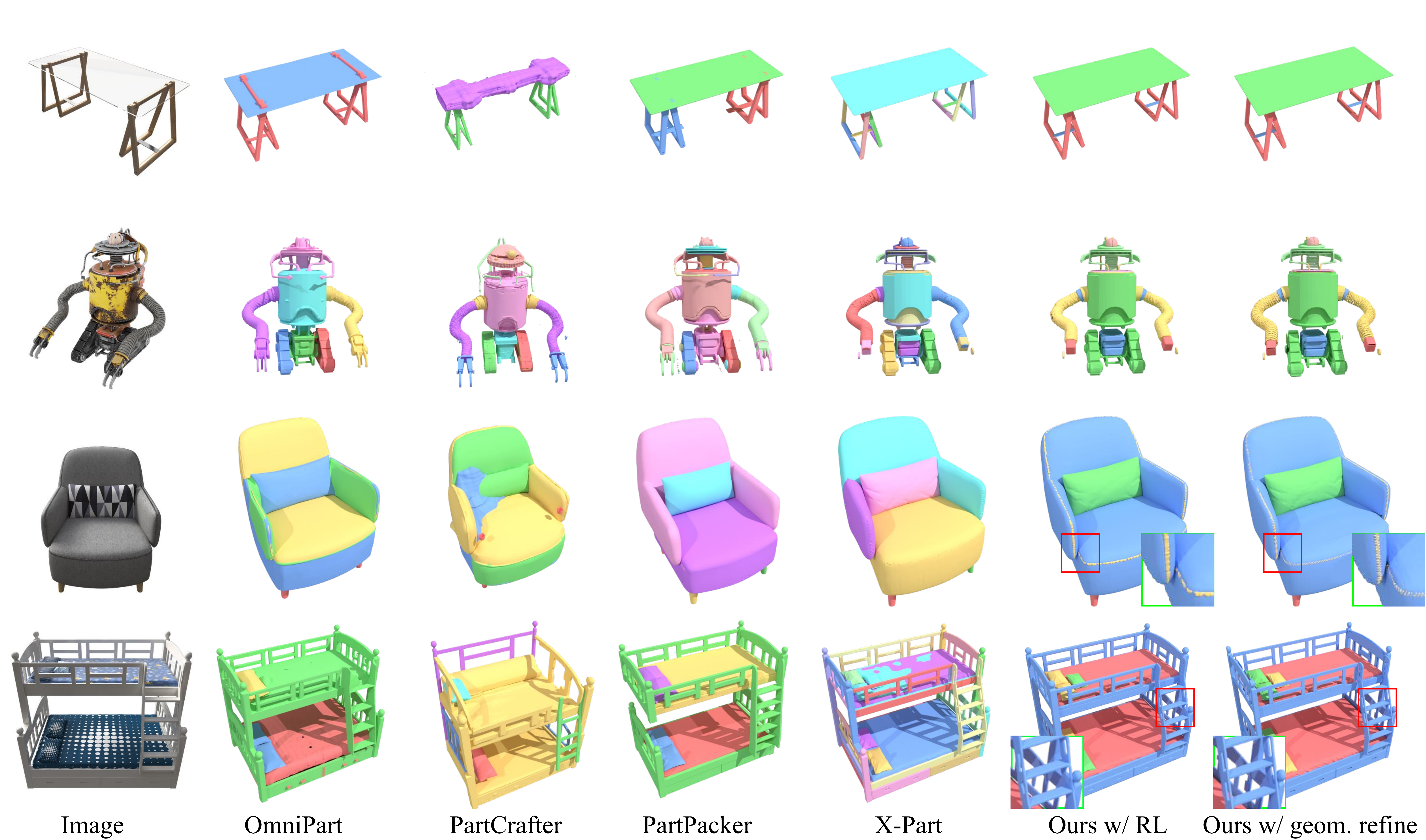}
\caption{Image-conditioned material-aware part generation comparison. Our method generates components with higher geometric quality, better material-region correspondence, and editable component IDs.}
\label{fig:dit_results}
\end{figure*}


\begin{table*}[t]
\centering

{
\small
\setlength{\tabcolsep}{3pt}
\begin{tabular}{lccrcccl}
\toprule
\multirow{2}{*}{Method} &
\multirow{2}{*}{CD ($\times\!10^{4}$)\,$\downarrow$} &
\multirow{2}{*}{F1@0.01\,$\uparrow$} &
\multirow{2}{*}{Avg.\ Tokens} &
\multicolumn{3}{c}{Enc-Dec Time (s)\,$\downarrow$} &
\multirow{2}{*}{Material Support} \\
\cmidrule(lr){5-7}
 & & & & 1 comp. & 16 comps. & 32 comps. & \\
\midrule
Hunyuan2.1                           & 2.179 & 75.8 & 4,096  & 3.572 & -- & -- & $\times$ \\
TRELLIS2~\cite{xiang2026native}    & 1.690 & 73.7 & 2,149  & \textbf{0.154} & -- & -- & $\times$ \\
PartCrafter~\cite{lin2026partcrafter} & 2.340 & 74.4 & 49152 & 1.782 & 12.958 & 26.508 & $\checkmark$ per-component \\
PartPacker~\cite{partpacked}         & 2.270 & 74.7 & 8,192  & 13.674 & 10.617 & 11.628 & \textdagger\ spatial \\
\midrule
\textbf{PartVAE (Ours)} & 2.140 & 75.2 & 4096 & 1.860 & \textbf{1.302} & \textbf{1.698} & $\checkmark$ material \\
\bottomrule
\end{tabular}
} 

\caption{VAE reconstruction and decode latency at $512^3$ resolution (10K sample points, F1 threshold 0.01). Decode time is measured for 1, 16, and 32 material components on the same GPU. Lower CD / decode time is better; higher F1 is better. Material support: $\times$ = none, \textdagger = spatial only, $\checkmark$ = material-editable components. Best results are marked in bold font.}
\label{tab:vae_reconstruction}

\end{table*}

\begin{figure}[t]
\centering
\includegraphics[width=\linewidth]{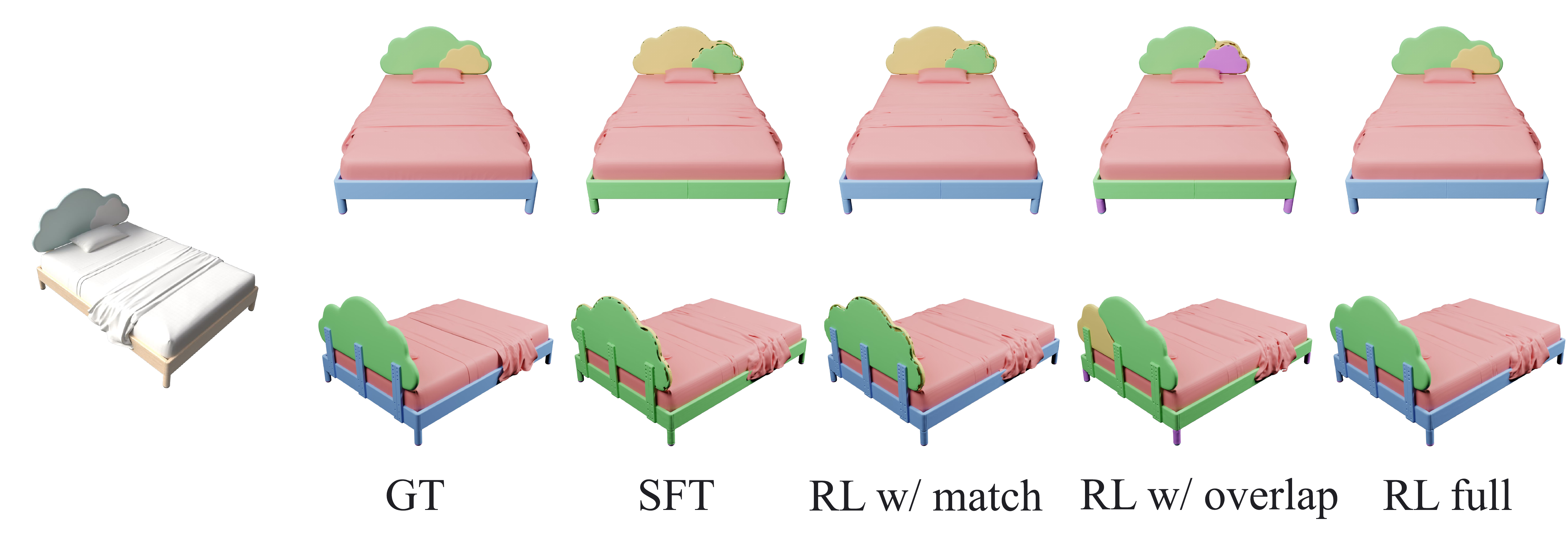}
\caption{Qualitative ablation of RL alignment. We compare the supervised fine-tuning (SFT) against different RL configurations. The matching reward improves component correspondence, while the overlap penalty reduces spatial conflicts; combining both yields the decomposition closest to the ground truth.}
\label{fig:rl_ablation}
\end{figure}

\subsection{Results}

\subsubsection{Material-Aware VAE Reconstruction}
\label{sec:vae_results}

Table~\ref{tab:vae_reconstruction} compares PartVAE against single-object and part-aware VAE baselines at $512^3$ resolution. Although TRELLIS2 achieves the lowest CD, and Hunyuan2.1 obtains the highest F1@0.01, neither of them provide material-editable component channels. PartCrafter preserves component identity but decodes each component independently, causing latency to grow with component count. PartPacker reduces this scaling cost with packed geometry, but uses twice the token budget and does not maintain one editable material slot per decoded channel. In contrast, PartVAE maintains a compact token budget of 4,096 and exhibits nearly constant decode latency as the component count scales to 32, all while delivering editable material parts and achieving geometric reconstruction competitive with these baselines.

Figure~\ref{fig:vae_results} visualizes reconstruction results. PartVAE directly decodes part-specific SDF fields with material-component IDs and obtains reconstruction quality comparable to Hunyuan2.1. Holistic VAEs do not expose editable regions, and per-component decoding baselines show higher latency cost as component count grows.

\subsubsection{Image-to-3D Material-Aware Generation}
\label{sec:dit_results}

Leveraging the compact latent space of PartVAE, we train an efficient downstream generative model that jointly synthesizes all parts within this unified representation, enabling the decoding of all components in a single forward pass.
Qualitative results are shown in Figure~\ref{fig:dit_results}. Compared to prior methods, our RL alignment enables PartMat to produce sharper material boundaries and more accurate structural correspondences between the reference image and the 3D output. Meanwhile, the geometry refinement network further enhances fine geometric details. Throughout this process, PartMat natively exposes editable component IDs for downstream material assignment. These visual advantages are quantitatively corroborated in Table~\ref{tab:dit_comparison}.


\begin{table}[t]
\centering

{
\small
\setlength{\tabcolsep}{3pt}
\begin{tabular}{lcc}
\toprule
Configuration & CD($\times\!10^{4}$)\,$\downarrow$ & F1@0.01 $\uparrow$ \\
\midrule
Local-box           & 3.03 & 71.8 \\
Packed-sem                     & 2.87 & 72.4 \\
\textbf{Global multi-channel (Ours)}                 & \textbf{2.53} & \textbf{74.3} \\
\bottomrule
\end{tabular}
} 

\caption{VAE architecture ablation for improving geometry. All variants use one global latent and up to $K{=}32$ material-component channels. Best results are marked in bold font.}
\label{tab:ablation_efficiency}

\end{table}


\subsubsection{Ablation Studies}
\label{sec:ablation}

To rigorously validate our architectural choices, we conduct comprehensive ablations, deferring further algorithmic and implementation details to the supplementary material. 

\begin{figure}[t]
\centering
\includegraphics[width=\linewidth]{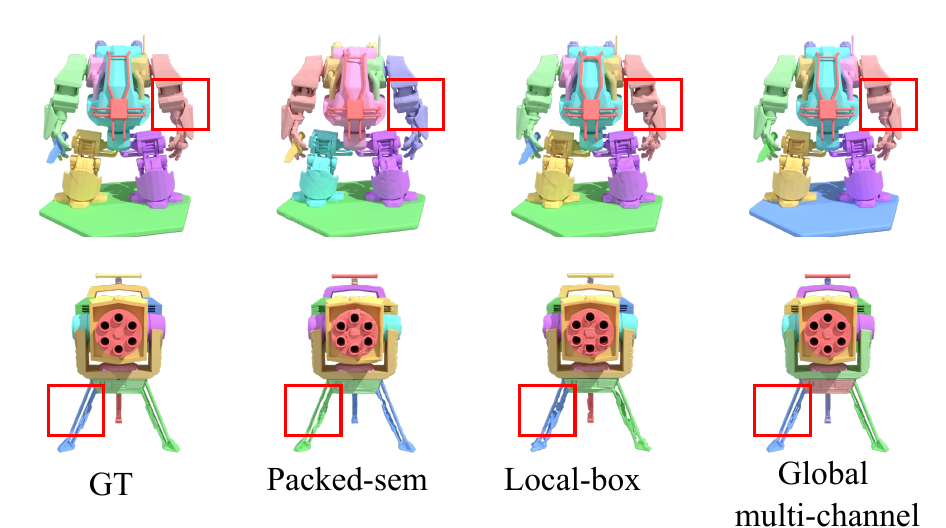}
\caption{Qualitative comparison of VAE representation paradigms. Unlike the missing thin structures in Local-box and jagged boundaries in Packed-sem, our Global multi-channel strategy achieves complete geometry and strict material disentanglement.}
\label{fig:vae_abla}
\end{figure}

First, we evaluate our VAE representation paradigm against two alternatives, with quantitative and qualitative results detailed in Table~\ref{tab:ablation_efficiency} and Figure~\ref{fig:vae_abla}. The \textit{Local-box} approach decodes component SDFs within normalized local spaces and explicitly predicts their global assembly poses. However, this coupled pose-geometry optimization introduces severe training difficulty, particularly struggling to reconstruct thin structures. Alternatively, the \textit{Packed-sem} strategy compresses components into a few spatially non-overlapping groups and relies on an auxiliary semantic classifier to disentangle them. This secondary classification frequently produces jagged material boundaries and destroys the continuous latent space strictly required by the downstream PartDiT. In contrast, our global multi-channel strategy natively models seamless junctions and guarantees strict material disentanglement without secondary classification noise.

Next, we further explore the specific contributions of our generative and refinement stages via the ablation experiments detailed in Table~\ref{tab:dit_comparison}. As visualized in Figure~\ref{fig:rl_ablation}, our direct-gradient RL alignment explicitly penalizes overlapping regions, effectively mitigating the blurred boundaries and spatial conflicts caused by purely supervised flow-matching. This facilitates more precise delineations between material regions. Finally, our sparse latent flow refinement restores high-frequency details lost in the compact global latent. This recovers sharp edges and micro structures crucial for photorealistic downstream applications, as visualized in Figure~\ref{fig:dit_results}.


\section{Conclusion}
\label{sec:conclusion}

We proposed \textbf{PartMat}, an efficient framework for material-aware 3D part generation. Integrating a global PartVAE, an RL-aligned PartDiT for overlap suppression, and a sparse flow refiner, PartMat significantly outperforms existing baselines. It uniquely maintains constant decoding latency regardless of the part count while natively generating disentangled, high-fidelity parts for seamless downstream PBR material workflows.

\noindent\textbf{Limitations.} PartMat assumes a fixed component capacity ($K{=}32$) and relies on high-quality material annotations, where weakly supervised alternatives may introduce boundary noise. Reconstructing tiny details remains challenging, requiring stronger refinement in future work.

\bibliography{egbib}

@inproceedings{liu2025partfield,
  title={Partfield: Learning 3d feature fields for part segmentation and beyond},
  author={Liu, Minghua and Uy, Mikaela Angelina and Xiang, Donglai and Su, Hao and Fidler, Sanja and Sharp, Nicholas and Gao, Jun},
  booktitle={Proceedings of the IEEE/CVF International Conference on Computer Vision},
  pages={9704--9715},
  year={2025}
}

@inproceedings{yan2026x,
  title={X-Part: High Fidelity And Structure Coherent Shape Decomposition And Completion},
  author={Yan, Xinhao and Xu, Jiachen and Li, Yang and Ma, Changfeng and Yang, Yunhan and Wang, Chunshi and Zhao, Zibo and Lai, Zeqiang and Zhao, Yunfei and Chen, Zhuo and others},
  booktitle={Proceedings of the IEEE/CVF Conference on Computer Vision and Pattern Recognition},
  pages={27062--27071},
  year={2026}
}

@article{zhang2025bang,
  title={BANG: Dividing 3D assets via generative exploded dynamics},
  author={Zhang, Longwen and Zhang, Qixuan and Jiang, Haoran and Bai, Yinuo and Yang, Wei and Xu, Lan and Yu, Jingyi},
  journal={ACM Transactions on Graphics (TOG)},
  volume={44},
  number={4},
  pages={1--21},
  year={2025},
  publisher={ACM New York, NY, USA}
}

@article{lin2026partcrafter,
  title={Partcrafter: Structured 3d mesh generation via compositional latent diffusion transformers},
  author={Lin, Yuchen and Lin, Chenguo and Pan, Panwang and Yan, Honglei and Yiqiang, Feng and Mu, Yadong and Fragkiadaki, Katerina},
  journal={Advances in neural information processing systems},
  volume={38},
  pages={35387--35415},
  year={2025}
}

@inproceedings{omnipart,
  title={Omnipart: Part-aware 3d generation with semantic decoupling and structural cohesion},
  author={Yang, Yunhan and Zhou, Yufan and Guo, Yuan-Chen and Zou, Zi-Xin and Huang, Yukun and Liu, Ying-Tian and Xu, Hao and Liang, Ding and Cao, Yan-Pei and Liu, Xihui},
  booktitle={Proceedings of the SIGGRAPH Asia 2025 Conference Papers},
  pages={1--12},
  year={2025}
}

@article{ding2025fullpart,
  title={FullPart: Generating each 3D Part at Full Resolution},
  author={Ding, Lihe and Dong, Shaocong and Li, Yaokun and Gao, Chenjian and Chen, Xiao and Han, Rui and Kuang, Yihao and Zhang, Hong and Huang, Bo and Huang, Zhanpeng and others},
  journal={arXiv preprint arXiv:2510.26140},
  year={2025}
}

@article{zhang20233dshape2vecset,
  title={3dshape2vecset: A 3d shape representation for neural fields and generative diffusion models},
  author={Zhang, Biao and Tang, Jiapeng and Niessner, Matthias and Wonka, Peter},
  journal={ACM Transactions On Graphics (TOG)},
  volume={42},
  number={4},
  pages={1--16},
  year={2023},
  publisher={ACM New York, NY, USA}
}

@inproceedings{he2026unipart,
  title={Unipart: Part-level 3d generation with unified 3d geom-seg latents},
  author={He, Xufan and Wu, Yushuang and Guo, Xiaoyang and Ye, Chongjie and Zhou, Jiaqing and Hu, Tianlei and Han, Xiaoguang and Du, Dong},
  booktitle={Proceedings of the IEEE/CVF Conference on Computer Vision and Pattern Recognition},
  pages={34227--34236},
  year={2026}
}

@inproceedings{qi2017pointnet,
  title={Pointnet: Deep learning on point sets for 3d classification and segmentation},
  author={Qi, Charles R and Su, Hao and Mo, Kaichun and Guibas, Leonidas J},
  booktitle={Proceedings of the IEEE conference on computer vision and pattern recognition},
  pages={652--660},
  year={2017}
}

@inproceedings{qi2017pointnetplusplus,
  title={Pointnet++: Deep hierarchical feature learning on point sets in a metric space},
  author={Qi, Charles Ruizhongtai and Yi, Li and Su, Hao and Guibas, Leonidas J},
  booktitle={Advances in neural information processing systems},
  volume={30},
  year={2017}
}

@inproceedings{mo2019partnet,
  title={Partnet: A large-scale benchmark for fine-grained and hierarchical part-level 3d object understanding},
  author={Mo, Kaichun and Zhu, Shilin and Chang, Angel X and Yi, Li and Tripathi, Subarna and Guibas, Leonidas J and Su, Hao},
  booktitle={Proceedings of the IEEE/CVF conference on computer vision and pattern recognition},
  pages={909--918},
  year={2019}
}

@article{wang2019dynamic,
  title={Dynamic graph cnn for learning on point clouds},
  author={Wang, Yue and Sun, Yongbin and Liu, Ziwei and Sarma, Sanjay E and Bronstein, Michael M and Solomon, Justin M},
  journal={ACM Transactions on Graphics (tog)},
  volume={38},
  number={5},
  pages={1--12},
  year={2019}
}

@inproceedings{golovinskiy2008randomized,
  title={Randomized cuts for 3D mesh analysis},
  author={Golovinskiy, Aleksey and Funkhouser, Thomas},
  booktitle={ACM SIGGRAPH Asia 2008 papers},
  pages={1--12},
  year={2008}
}

@article{shapira2008consistent,
  title={Consistent mesh partitioning and skeletonisation using the shape diameter function},
  author={Shapira, Lior and Shamir, Ariel and Cohen-Or, Daniel},
  journal={The Visual Computer},
  volume={24},
  pages={249--259},
  year={2008}
}

@inproceedings{chen2019baenet,
  title={BAE-NET: Branched autoencoder for shape co-segmentation},
  author={Chen, Zhiqin and Yin, Kangxue and Fisher, Matthew and Chaudhuri, Siddhartha and Zhang, Hao},
  booktitle={Proceedings of the IEEE/CVF International Conference on Computer Vision},
  pages={8490--8499},
  year={2019}
}

@article{zhao2025hunyuan3d,
  title={Hunyuan3d 2.0: Scaling diffusion models for high resolution textured 3d assets generation},
  author={Zhao, Zibo and Lai, Zeqiang and Lin, Qingxiang and Zhao, Yunfei and Liu, Haolin and Yang, Shuhui and Feng, Yifei and Yang, Mingxin and Zhang, Sheng and Yang, Xianghui and others},
  journal={arXiv preprint arXiv:2501.12202},
  year={2025}
}

@article{hunyuan3d21,
  title={Hunyuan3d 2.1: From images to high-fidelity 3d assets with production-ready pbr material},
  author={Hunyuan3D, Team and Yang, Shuhui and Yang, Mingxin and Feng, Yifei and Huang, Xin and Zhang, Sheng and He, Zebin and Luo, Di and Liu, Haolin and Zhao, Yunfei and others},
  journal={arXiv preprint arXiv:2506.15442},
  year={2025}
}

@article{liu2026flow,
  title={Flow-grpo: Training flow matching models via online rl},
  author={Liu, Jie and Liu, Gongye and Liang, Jiajun and Li, Yangguang and Liu, Jiaheng and Wang, Xintao and Wan, Pengfei and Zhang, Di and Ouyang, Wanli},
  journal={Advances in neural information processing systems},
  volume={38},
  pages={40783--40818},
  year={2026}
}

@article{partpacked,
  title={Efficient part-level 3d object generation via dual volume packing},
  author={Tang, Jiaxiang and Lu, Ruijie and Li, Max and Hao, Zekun and Li, Xuan and Wei, Fangyin and Song, Shuran and Zeng, Gang and Liu, Ming-Yu and Lin, Tsung-Yi},
  journal={Advances in Neural Information Processing Systems},
  volume={38},
  pages={27115--27137},
  year={2025}
}

@inproceedings{xiang2025structured,
  title={Structured 3d latents for scalable and versatile 3d generation},
  author={Xiang, Jianfeng and Lv, Zelong and Xu, Sicheng and Deng, Yu and Wang, Ruicheng and Zhang, Bowen and Chen, Dong and Tong, Xin and Yang, Jiaolong},
  booktitle={Proceedings of the IEEE/CVF conference on computer vision and pattern recognition},
  pages={21469--21480},
  year={2025}
}

@article{oquab2023dinov2,
  title={{DINOv2}: Learning Robust Visual Features without Supervision},
  author={Oquab, Maxime and Darcet, Timoth{\'e}e and Moutakanni, Th{\'e}o and Vo, Huy V and Szafraniec, Marc and Khalidov, Vasil and Fernandez, Pierre and Haziza, Daniel and Massa, Francisco and El-Nouby, Alaaeldin and others},
  journal={Transactions on Machine Learning Research},
  year={2024}
}

@inproceedings{chen2025partgen,
  title={Partgen: Part-level 3d generation and reconstruction with multi-view diffusion models},
  author={Chen, Minghao and Shapovalov, Roman and Laina, Iro and Monnier, Tom and Wang, Jianyuan and Novotny, David and Vedaldi, Andrea},
  booktitle={Proceedings of the Computer Vision and Pattern Recognition Conference},
  pages={5881--5892},
  year={2025}
}

@inproceedings{tulsiani2017primitives,
  title={Learning shape abstractions by assembling volumetric primitives},
  author={Tulsiani, Shubham and Su, Hao and Guibas, Leonidas J and Efros, Alexei A and Malik, Jitendra},
  booktitle={Proceedings of the IEEE conference on computer vision and pattern recognition},
  pages={2635--2643},
  year={2017}
}

@article{li2017grass,
  title={{GRASS}: Generative Recursive Autoencoders for Shape Structures},
  author={Li, Jun and Xu, Kai and Chaudhuri, Siddhartha and Yumer, Ersin and Zhang, Hao and Guibas, Leonidas},
  journal={ACM Transactions on Graphics},
  volume={36},
  number={4},
  year={2017}
}

@inproceedings{lorensen1987marching,
  title={Marching cubes: A high resolution 3D surface construction algorithm},
  author={Lorensen, William E and Cline, Harvey E},
  booktitle={Proceedings of the 14th Annual Conference on Computer Graphics and Interactive Techniques (SIGGRAPH)},
  pages={163--169},
  year={1987}
}

@inproceedings{ren2024xcube,
  title={Xcube: Large-scale 3d generative modeling using sparse voxel hierarchies},
  author={Ren, Xuanchi and Huang, Jiahui and Zeng, Xiaohui and Museth, Ken and Fidler, Sanja and Williams, Francis},
  booktitle={Proceedings of the IEEE/CVF conference on computer vision and pattern recognition},
  pages={4209--4219},
  year={2024}
}

@inproceedings{hong2024lrm,
  title={Lrm: Large reconstruction model for single image to 3d},
  author={Hong, Yicong and Zhang, Kai and Gu, Jiuxiang and Bi, Sai and Zhou, Yang and Liu, Difan and Liu, Feng and Sunkavalli, Kalyan and Bui, Trung and Tan, Hao},
  booktitle={International Conference on Learning Representations},
  volume={2024},
  pages={50678--50702},
  year={2024}
}

@article{li2025triposg,
  title={Triposg: High-fidelity 3d shape synthesis using large-scale rectified flow models},
  author={Li, Yangguang and Zou, Zi-Xin and Liu, Zexiang and Wang, Dehu and Liang, Yuan and Yu, Zhipeng and Liu, Xingchao and Guo, Yuan-Chen and Liang, Ding and Ouyang, Wanli and others},
  journal={IEEE Transactions on Pattern Analysis and Machine Intelligence},
  year={2025},
  publisher={IEEE}
}

@article{wu2026direct3d,
  title={Direct3d-s2: Gigascale 3d generation made easy with spatial sparse attention},
  author={Wu, Shuang and Lin, Youtian and Zhang, Feihu and Zeng, Yifei and Yang, Yikang and Bao, Yajie and Qian, Jiachen and Zhu, Siyu and Cao, Xun and Torr, Philip H. S. and Yao, Yao},
  journal={Advances in Neural Information Processing Systems},
  volume={38},
  pages={170778--170804},
  year={2025}
}

@inproceedings{liu2024part123,
  title={Part123: Part-aware 3D reconstruction from a single-view image},
  author={Liu, Anran and Lin, Cheng and Liu, Yuan and Long, Xiaoxiao and Dou, Zhiyang and Guo, Hao-Xiang and Luo, Ping and Wang, Wenping},
  booktitle={ACM SIGGRAPH 2024 Conference Papers},
  pages={1--12},
  year={2024}
}

@article{yang2025holopart,
  title={HoloPart: Generative 3D Part Amodal Segmentation},
  author={Yang, Yunhan and Guo, Yuan-Chen and Huang, Yukun and Zou, Zi-Xin and Yu, Zhipeng and Li, Yangguang and Cao, Yan-Pei and Liu, Xihui},
  journal={arXiv preprint arXiv:2504.07943},
  year={2025}
}

@inproceedings{dora,
  title={Dora: Sampling and benchmarking for 3d shape variational auto-encoders},
  author={Chen, Rui and Zhang, Jianfeng and Liang, Yixun and Luo, Guan and Li, Weiyu and Liu, Jiarui and Li, Xiu and Long, Xiaoxiao and Feng, Jiashi and Tan, Ping},
  booktitle={Proceedings of the IEEE/CVF Conference on Computer Vision and Pattern Recognition},
  pages={16251--16261},
  year={2025}
}

@inproceedings{xiang2026native,
  title={Native and compact structured latents for 3d generation},
  author={Xiang, Jianfeng and Chen, Xiaoxue and Xu, Sicheng and Wang, Ruicheng and Lv, Zelong and Deng, Yu and Zhu, Hongyuan and Dong, Yue and Zhao, Hao and Yuan, Nicholas Jing and others},
  booktitle={Proceedings of the IEEE/CVF Conference on Computer Vision and Pattern Recognition},
  pages={14419--14429},
  year={2026}
}

@inproceedings{lai2026lattice,
  title={Lattice: Democratize high-fidelity 3d generation at scale},
  author={Lai, Zeqiang and Zhao, Yunfei and Zhao, Zibo and Liu, Haolin and Lin, Qingxiang and Huang, Jingwei and Guo, Chunchao and Yue, Xiangyu},
  booktitle={Proceedings of the IEEE/CVF Conference on Computer Vision and Pattern Recognition},
  pages={19982--19992},
  year={2026}
}

@inproceedings{liang2026leapalign,
  title={LeapAlign: Post-Training Flow Matching Models at Any Generation Step by Building Two-Step Trajectories},
  author={Liang, Zhanhao and Yang, Tao and Wu, Jie and Feng, Chengjian and Zheng, Liang},
  booktitle={Proceedings of the IEEE/CVF Conference on Computer Vision and Pattern Recognition},
  pages={23238--23248},
  year={2026}
}

@inproceedings{zhao2021pointtransformer,
  title={Point transformer},
  author={Zhao, Hengshuang and Jiang, Li and Jia, Jiaya and Torr, Philip HS and Koltun, Vladlen},
  booktitle={Proceedings of the IEEE/CVF international conference on computer vision},
  pages={16259--16268},
  year={2021}
}

@inproceedings{kirillov2023segmentanything,
  title={Segment Anything},
  author={Kirillov, Alexander and Mintun, Eric and Ravi, Nikhila and Mao, Hanzi and Rolland, Chloe and Gustafson, Laura and Xiao, Tete and Whitehead, Spencer and Berg, Alexander C. and Lo, Wan-Yen and others},
  booktitle={Proceedings of the IEEE/CVF International Conference on Computer Vision},
  pages={4015--4026},
  year={2023}
}

@inproceedings{abdelreheem2023satr,
  title={SATR: Zero-Shot Semantic Segmentation of 3D Shapes},
  author={Abdelreheem, Ahmed and Skorokhodov, Ivan and Ovsjanikov, Maks and Wonka, Peter},
  booktitle={Proceedings of the IEEE/CVF International Conference on Computer Vision},
  pages={15166--15179},
  year={2023}
}

@inproceedings{liu2023partslip,
  title={PartSLIP: Low-Shot Part Segmentation for 3D Point Clouds via Pretrained Image-Language Models},
  author={Liu, Minghua and Zhu, Yinhao and Cai, Hong and Han, Shizhong and Ling, Zhan and Porikli, Fatih and Su, Hao},
  booktitle={Proceedings of the IEEE/CVF Conference on Computer Vision and Pattern Recognition},
  pages={21736--21746},
  year={2023}
}

@article{tang2024samesh,
  title={Segment Any Mesh: Zero-Shot Mesh Part Segmentation via Lifting Segment Anything 2 to 3D},
  author={Tang, George and Zhao, William and Ford, Logan and Benhaim, David and Zhang, Paul},
  journal={arXiv preprint arXiv:2408.13679},
  year={2024}
}

@inproceedings{thai2024threebytwo,
  title={3$\times$ 2: 3d object part segmentation by 2d semantic correspondences},
  author={Thai, Anh and Wang, Weiyao and Tang, Hao and Stojanov, Stefan and Rehg, James M and Feiszli, Matt},
  booktitle={European Conference on Computer Vision},
  pages={149--166},
  year={2024},
  organization={Springer}
}

@inproceedings{umam2024partdistill,
  title={PartDistill: 3D Shape Part Segmentation by Vision-Language Model Distillation},
  author={Umam, Ardian and Yang, Cheng-Kun and Chen, Min-Hung and Chuang, Jen-Hui and Lin, Yen-Yu},
  booktitle={Proceedings of the IEEE/CVF Conference on Computer Vision and Pattern Recognition},
  pages={3470--3479},
  year={2024}
}

@inproceedings{zhong2024meshsegmenter,
  title={MeshSegmenter: Zero-Shot Mesh Semantic Segmentation via Texture Synthesis},
  author={Zhong, Ziming and Xu, Yanyu and Li, Jing and Xu, Jiale and Li, Zhengxin and Yu, Chaohui and Gao, Shenghua},
  booktitle={European Conference on Computer Vision},
  pages={182--199},
  year={2024}
}

@article{yang2024sampart3d,
  title={SAMPart3D: Segment Any Part in 3D Objects},
  author={Yang, Yunhan and Huang, Yukun and Guo, Yuan-Chen and Lu, Liangjun and Wu, Xiaoyang and Lam, Edmund Y. and Cao, Yan-Pei and Liu, Xihui},
  journal={arXiv preprint arXiv:2411.07184},
  year={2024}
}

@article{ma2025p3sam,
  title={P3-SAM: Native 3D Part Segmentation},
  author={Ma, Changfeng and Li, Yang and Yan, Xinhao and Xu, Jiachen and Yang, Yunhan and Wang, Chunshi and Zhao, Zibo and Guo, Yanwen and Chen, Zhuo and Guo, Chunchao},
  journal={arXiv preprint arXiv:2509.06784},
  year={2025}
}

@inproceedings{ahmed2025kestrel,
  title={Kestrel: 3D Multimodal LLM for Part-Aware Grounded Description},
  author={Ahmed, Mahmoud and Fei, Junjie and Ding, Jian and Bakr, Eslam Mohamed and Elhoseiny, Mohamed},
  booktitle={Proceedings of the IEEE/CVF International Conference on Computer Vision},
  pages={8973--8983},
  year={2025}
}

@inproceedings{chen2025autopartgen,
  title={AutoPartGen: Autoregressive 3D Part Generation and Discovery},
  author={Chen, Minghao and Wang, Jianyuan and Shapovalov, Roman and Monnier, Tom and Jung, Hyunyoung and Wang, Dilin and Ranjan, Rakesh and Laina, Iro and Vedaldi, Andrea},
  booktitle={Advances in Neural Information Processing Systems},
  year={2025}
}

@inproceedings{shen2024spacemesh,
  title={{SpaceMesh}: A Continuous Representation for Learning Manifold Surface Meshes},
  author={Shen, Tianchang and Li, Zhaoshuo and Law, Marc and Atzmon, Matan and Fidler, Sanja and Lucas, James and Gao, Jun and Sharp, Nicholas},
  booktitle={SIGGRAPH Asia 2024 Conference Papers},
  pages={1--11},
  year={2024}
}

@inproceedings{li2026meshflow,
  title={{MeshFlow}: Efficient Artistic Mesh Generation via {MeshVAE} and Flow-based Diffusion Transformer},
  author={Li, Weiyu and Toisoul, Antoine and Monnier, Tom and Shapovalov, Roman and Ranjan, Rakesh and Tan, Ping and Vedaldi, Andrea},
  booktitle={Proceedings of the IEEE/CVF Conference on Computer Vision and Pattern Recognition},
  year={2026}
}

@article{wang2026nexus,
  title={Nexus: Native Mesh Generation with Diffusion},
  author={Wang, Hanxiao and Liu, Ying-Tian and Guo, Yuan-Chen and Feng, Qi-Yuan and Zou, Zi-Xin and Liang, Ding and Zhang, Biao and Cao, Yan-Pei},
  journal={ACM Transactions on Graphics (TOG)},
  volume={45},
  number={4},
  pages={1--14},
  year={2026},
  publisher={ACM New York, NY, USA}
}

@article{long2026lato2,
  title={{LATO.2}: Factorized 3D Mesh Generation with Vertex and Topology Flow},
  author={Long, Hang and Zhao, Tianhao and Lin, Junkai and Zhang, Youjia and Guo, Huipeng and Liang, Rendong and Xu, Jiale and Hladk{\'y}, Jozef and Nie{\ss}ner, Matthias and Hu, Yuanming and Yang, Wei},
  journal={arXiv preprint arXiv:2607.10623},
  year={2026}
}


\end{document}